%% file: main.tex
\documentclass[11pt]{article}

\PassOptionsToPackage{dvipsnames}{xcolor}

\usepackage[final]{acl}

\usepackage{times}
\usepackage{latexsym}
\usepackage[T1]{fontenc}
\usepackage[utf8]{inputenc}
\usepackage{microtype}
\usepackage{inconsolata}
\usepackage{graphicx}
\usepackage{booktabs}
\usepackage{amsmath}
\usepackage{amssymb}
\usepackage{xcolor}
\usepackage[normalem]{ulem}
\usepackage{url}
\usepackage{tikz}
\usetikzlibrary{positioning,arrows.meta,calc}
\usepackage{algorithm}
\usepackage{algpseudocode}
\usepackage{enumitem}
\usepackage{amsthm}
\usepackage{multirow}
\usepackage{listings}
\usepackage[most,breakable]{tcolorbox}
\theoremstyle{definition}
\newtheorem{definition}{Definition}

\newcommand{\crtypo}[1]{#1}          
\newcommand{\crpol}[1]{#1}           

\graphicspath{{figures/}}

\lstdefinestyle{prompt}{
  basicstyle=\ttfamily\scriptsize,
  breaklines=true,
  breakatwhitespace=false,
  columns=fullflexible,
  keepspaces=true,
  showstringspaces=false,
  frame=single,
  framerule=0.3pt,
  framesep=4pt,
  xleftmargin=0pt,
  xrightmargin=0pt,
  belowskip=0.5em,
  aboveskip=0.5em,
  literate={\`}{{`}}1
}

\title{MDForge: Agentic Molecular Dynamics Pipeline Design under\\ Sparse Simulator Feedback}

\author{
    \textbf{Zehong Wang}$^{1}$\;
    \textbf{Yijun Ma}$^{1}$\;
    \textbf{Connor R. Schmidt}$^{1}$\;
    \textbf{Tianyi Ma}$^{1}$\;
    \textbf{Weixiang Sun}$^{1}$\;
    \\
    \textbf{Ziming Li}$^{2}$\;
    \textbf{Xiaoguang Guo}$^{2}$\; 
    \textbf{Chuxu Zhang}$^{2}$\;
    \textbf{Matthew  J. Webber}$^{1}$\;
    \textbf{Yanfang Ye}$^{1,\dagger}$
    \\
    \textsuperscript{1} University of Notre Dame\; 
    \textsuperscript{2} University of Connecticut\; 
    \\
    $^\dagger$ Corresponding Author
    \\
    \texttt{<zwang43,yye7>@nd.edu}
}

\begin{document}
\maketitle

\begin{abstract}
\input{sections/abstract}
\end{abstract}

\input{sections/introduction}
\input{sections/related_work}
\input{sections/problem}

\input{sections/method}

\input{sections/experiments}
\input{sections/conclusion}

\input{sections/limitations}
\input{sections/ethics}

\bibliography{reference}

\newpage
\appendix
\input{appendix/related_work}
\input{appendix/method}
\input{appendix/experimental_details}
\input{appendix/additional_results}
\input{appendix/prompt}

\end{document}

%% file: sections/abstract.tex
Molecular dynamics (MD) is the canonical in-silico method for atomistic molecular science, simulating molecular behavior from first-principles physics. Designing an MD pipeline for a new system requires substantial expert knowledge: running it on even one molecule is expensive, ruling out trial-and-error. We automate this expert pipeline-design process with an LLM agent. Unlike existing MD agents that orchestrate a predefined tool set, we treat pipeline design as open-ended code generation in which the agent's behavior is reshaped online by verbal reward. Specifically, we build MDForge, an LLM agent whose in-context update rule densifies the sparse reward via a multi-agent debate among physics experts. On three SAMPL host–guest binding free-energy benchmarks (CB[7], OAH, CBClip), MDForge automatically designs MD pipelines competitive with human experts. Deployed on a library of unseen candidate guests, its CB[7] pipeline discovers a novel binder that wet-lab competition NMR confirms is a high-affinity, picomolar CB[7] binder ($K_a \approx 8 \times 10^{12}$~M$^{-1}$). Our data and code are available at \url{https://github.com/Zehong-Wang/MDForge}.

%% file: sections/introduction.tex
\section{Introduction}
\label{sec:intro}

\begin{figure}[!t]
  \centering
  \includegraphics[width=\linewidth]{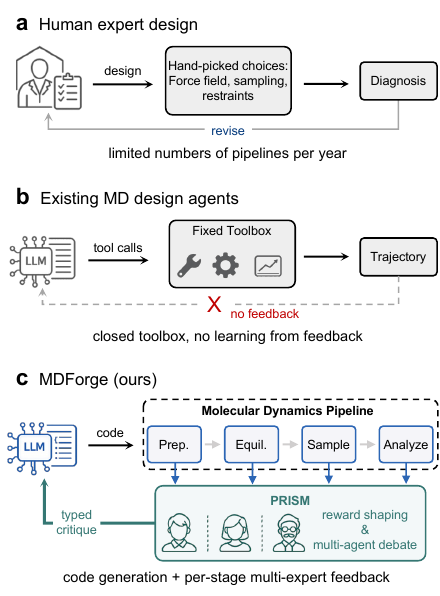}
  \caption{\textbf{Three paradigms for MD pipeline design.} (a) A human expert hand-picks each stage and iteratively revises. (b) Existing LLM agents for MD design call a fixed MD toolbox with no run-time feedback. (c) MDForge emits the pipeline as code and refines it via PRISM, a multi-expert debate over per-stage diagnostics that returns a typed critique.}
  \label{fig:introduction}
\end{figure} 

Molecular dynamics (MD) simulation has long been the canonical in-silico method for studying molecular behavior at atomistic resolution~\citep{karplus2002md,hollingsworth2018md}. By integrating \crtypo{first-principles} equations of motion, MD produces atomistic trajectories from which a researcher can understand binding affinities, conformational ensembles, reaction pathways, and material properties. Several of these quantities are accessible to wet-lab measurement only at considerable expense and time; others, such as the transient conformational states populated during an enzymatic catalytic cycle, are not directly observable at all. These properties have made MD a mainstay of biology, drug discovery, and chemistry for decades~\citep{behler2021nnp,unke2021mlff}.

Designing an MD pipeline for a new molecular system typically requires the work of trained scientists, and the throughput of new pipelines is correspondingly modest~\citep{mey2020bestpractices}. A free-energy calculation illustrates this: it involves joint specification of a binding-pose hypothesis, force-field parameterization, equilibration schedule, sampling protocol, restraints, and an estimator. These choices interact non-trivially and few are universal: a pipeline tuned for one system class rarely transfers, because the dominant physics differs across system types~\citep{mobley2017predicting,schindler2020largescale}. 
The recent surge of AI-driven molecular predictors does not remove this need. A data-driven predictor outputs a target property value (\textit{e.g.}, a binding affinity)~\citep{merchant2023gnome,ross2022molformer,wang2026molrep,ye2026latentchem} but does not produce the atomistic trajectory MD does, so it cannot supply the mechanistic account that physics-based simulation is invoked for in the first place. Its applicability is also bounded by the chemical space it was trained on: the model has no foothold on a system class for which no large labeled corpus exists~\citep{wu2018moleculenet,yang2019analyzing}, and on inputs outside its training distribution its predictions degrade silently~\citep{bender2021ai,vantilborg2022activitycliffs}. MD therefore remains indispensable for mechanistic understanding~\citep{bottaro2018biophysical}, but designing its pipeline for a new system is an expert task. 

In this work, we aim to design an agentic AI system that can automate the MD design by replicating the work of a trained expert. Faced with a new molecular system, the expert~\citep{cournia2017relative} first inspects its chemistry, charges, rigidity, and binding mode, and these observations dictate every downstream choice: the force-field family, the equilibration schedule, the sampling protocol, the restraint scheme, and the estimator. The pipeline is then run; the expert reads the diagnostics it returns (divergence traces, free-energy convergence plots, restraint-release artifacts), identifies which subsystem misbehaved, and revises the pipeline for the next trial. Several recent LLM agents target this automation. For example, MDCrow~\citep{mdcrow2024} wraps a general-purpose MD toolset (force-field setup, simulation, trajectory analysis) in LangChain-style tool calls~\citep{yao2023react}; MDAgent~\citep{ma2026mdagent} extends the pattern with a memory module that reuses parameter choices and analytical logic from prior tasks~\citep{zhao2024expel,chen2024automanual}; DynaMate~\citep{dynamate2025} ports the same tool-calling pattern to binding free-energy workflows. Yet none of these systems matches exactly what an MD expert does. Their tool-calling resembles the expert's selection of pipeline pieces, but only from a fixed toolbox, narrowing what the expert can otherwise compose. Likewise, none of them uses the feedback the workflow returns, yet that feedback \crtypo{(though sparse)} is what the expert depends on to refine the pipeline.

To tackle both gaps, we propose \textbf{MDForge}, an LLM-driven agent that frames MD pipeline design as open-ended code generation~\citep{wang2023voyager} under verbal reinforcement learning~\citep{shinn2023reflexion}. Code generation matches the expert's actual action space, which is not a preregistered toolbox but whatever the new system asks for. Verbal RL matches the expert's iteration habit, where each trial's diagnostics drive the next pipeline. This framing surfaces the central technical challenge of the paper: building an agent that can learn from very few feedback signals. Each signal arrives only after a full MD workflow run, whose GPU-hour cost confines each task to a small trial budget, far too limited for the agent to iteratively update behaviors in a typical approach~\citep{wang2026planning,chen2026longhorizon,gupta2025sparse}.

At the heart of MDForge is Process-Reward Interpretation via Subsystem Mediation \textbf{(PRISM)}, an in-context update rule that turns the handful of terminal rewards into a dense, typed learning signal along two axes. First, PRISM exploits the staged nature of an MD pipeline (preparation, equilibration, production sampling, analysis): it harvests per-stage diagnostics from the simulator's intermediate outputs, so the agent receives feedback at every stage boundary rather than only at the end of the run~\citep{lightman2023verify,uesato2022process,wang2024mathshepherd}. Second, PRISM launches a panel of physics experts (force field, sampling, analysis) to debate each diagnostic~\citep{du2024debate} and produce a typed, subsystem-attributable critique that reshapes MDForge's behavior, surfacing the kind of physical interpretation only experts can provide. Empirically, MDForge produces pipelines comparable to expert hand-designs on three SAMPL host–guest binding free-energy benchmarks~\citep{sampl4cb7,sampl5cbclip} (CB[7], OAH, and CBClip). The best AI-designed CB[7] pipeline, applied to a library of unseen candidate guests, discovers a novel binder confirmed by wet-lab competition NMR to be a  high-affinity (picomolar) CB[7] binder ($K_a \approx 8 \times 10^{12}$~M$^{-1}$).

%% file: sections/related_work.tex
\section{Related Work}
\label{sec:related}

\paragraph{Molecular dynamics.}
MDForge sits atop an established physics-based MD stack rather than competing with any of its parts: alchemical FEP/TI with BAR/MBAR estimators~\citep{bennett1976bar,shirts2008mbar,mey2020bestpractices}, mature simulation engines~\citep{eastman2017openmm,abraham2015gromacs,case2023amber}, and standard biomolecular force-field families. Recent neural work replaces individual slices of this stack with learned components: ML force fields~\citep{behler2021nnp,unke2021mlff}, structure predictors~\citep{jumper2021alphafold}, and equilibrium samplers~\citep{noe2019boltzmann}. MDForge automates the \emph{workflow} itself as executable code, so the design space is a program-synthesis over the existing toolset rather than the parameter space of a fixed pipeline template.

\paragraph{Autonomous science agents.}
LLM-driven scientific agents have integrated literature search, hypothesis proposal, and code synthesis into runnable discovery pipelines across chemistry, materials, and biology~\citep{ye2025llms4all,boiko2023coscientist,bran2024chemcrow,sakana2024aiscientist,merchant2023gnome}. MD-specific agents have converged on a \emph{tool-calling} pattern that orchestrates a fixed library of engines, force fields, and analysis routines under LLM control~\citep{mdcrow2024,ma2026mdagent,dynamate2025,chandrasekhar2025namdagent,shi2025mdagent}. MDForge instead treats MD pipeline design as open-ended code generation in the lineage of program-synthesis agents~\citep{wang2023voyager,romera2024funsearch}, operating in a regime where the supervisory signal is both sparse (one terminal reward per trial) and expensive (GPU-hours of MD execution).

See Appendix~\ref{app:related-extended} for extended discussion.

%% file: sections/problem.tex
\section{Problem Setup}
\label{sec:problem}

\paragraph{Task.}
Given a target system class $\mathcal{T} = \{s_1, \ldots, s_M\}$ of related molecular systems with experimental references $\{y_{\exp}(s_m)\}$ for some target observable $y$ (e.g., binding free-energy), the agent emits an executable MD pipeline $\pi \in \Pi$ that, applied across $\mathcal{T}$, minimizes the mean per-system prediction error $\mathcal{L}(\pi) = \tfrac{1}{M}\sum_m |\hat{y}_\pi(s_m) - y_{\exp}(s_m)|$. 

\paragraph{POMDP.}
We cast the design loop as $\mathcal{M} = (\mathcal{S}, \mathcal{A}, \mathcal{O}, T, R, \gamma)$: state $\mathcal{S}$ is the design history $(\pi_{1:t}, D_{\pi_{1:t}})$ of pipelines tried and their stage-level diagnostics; the action space $\mathcal{A} = \Pi$ is the open-ended space of executable programs that emit an MD workflow over four canonical stages (Prep, Equilibration, Production, Analysis); observations $\mathcal{O} \subseteq \mathcal{V}$ are the natural-language documents the simulator returns; transitions $T$ are deterministic, governed by physics and the toolchain; the reward $R$ realizes only at horizon as $r^*_\pi = -\mathcal{L}(\pi)$; and $\gamma=1$. The reward is therefore both \emph{sparse} (one event per pipeline) and \emph{expensive} (a GPU-hour production run per trial). Therefore, we intend to use verbal RL to solve the problem. 

\begin{definition}[Verbal RL]
\label{def:verbal-rl}
With $\mathcal{V}$ the space of natural-language strings, a POMDP is \emph{verbal} if $\mathcal{A}, \mathcal{O} \subseteq \mathcal{V}$ and the policy is an LLM with frozen parameters $\theta$ acting on a textual context $\mathcal{C}_t \in \mathcal{V}$,
\begin{align}
    \pi_{t+1} &\,\sim\, \mathrm{LLM}_\theta(\,\cdot\,\mid\,\mathcal{C}_{t+1}), \\
    \mathcal{C}_{t+1} &\,=\, \mathrm{Update}(\mathcal{C}_t,\, \pi_t,\, o_t,\, r_t),
\end{align}
where $\mathrm{Update}$ is an LLM call folding each trial outcome $(o_t, r_t)$ back into the context.
\end{definition}

%% file: sections/method.tex
\section{MDForge}
\label{sec:method}

We present MDForge, the LLM agent that instantiates the verbal RL for automatic molecular dynamics workflow design. The framework is shown in Figure~\ref{fig:framework} with the full protocol in Appendix~\ref{app:method-details}.

\begin{figure*}[!ht]
  \centering
  \includegraphics[width=\linewidth]{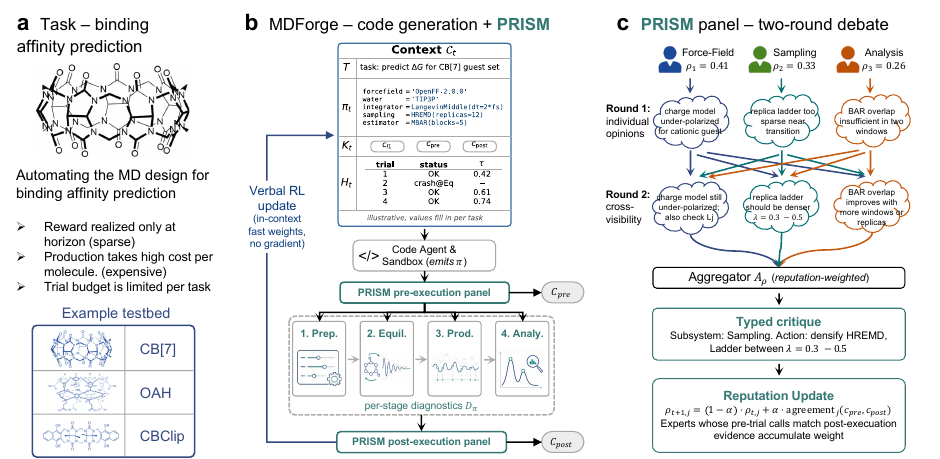}
  \caption{\textbf{Overview of MDForge.} (a) Automating MD design for binding affinity prediction, instantiated on the SAMPL CB[7], OAH, and CBClip testbeds. (b) A Code agent reads the context bundle $\mathcal{C}_t{=}\{T, \pi_t, K_t, H_t\}$ (task, current pipeline as typed code, critique set, and headline-metric trial history) and emits an executable pipeline through a sandbox. Execution proceeds through $K{=}4$ canonical stages (Preparation, Equilibration, Production, Analysis), yielding per-stage diagnostics $D_\pi$. A PRISM panel reviews $\pi$ pre- and post-execution to emit typed critiques $c_{\text{pre}}, c_{\text{post}}$, which feed back into $\mathcal{C}_{t+1}$ as in-context fast-weight updates. (c) $J{=}3$ specialists (Force-Field, Sampling, Analysis) with reputations $\rho_j$ first produce independent opinions (Round~1), then revise under cross-visibility (Round~2); a reputation-weighted aggregator $\mathcal{A}_\rho$ emits a single typed critique (subsystem + action).}
  \vspace{-10pt}
  \label{fig:framework}
\end{figure*}

\subsection{Design Rationale}
\label{sec:method:rationale}

Verbal RL reduces trial-to-trial learning to the context rewrite $\mathcal{C}_t \mapsto \mathcal{C}_{t+1}$, equivalently a fast-weight update of the LLM's induced state without touching its parameters~\citep{schmidhuber1992fastweights,ba2016fastweights,schlag2021linear}. Under an expensive reward (each trial requires GPU-hours of MD execution before producing $r^*_\pi$), the only lever is to enrich the information each reward event carries. Two general approaches densify a sparse signal: (i) split the reward across the pipeline so each part receives its own signal~\citep{lightman2023verify}, and (ii) attach explanatory text to each signal value~\citep{shinn2023reflexion}. An MD pipeline supplies a natural instantiation of each: it is staged along an execution sequence, yielding per-stage diagnostics, and naturally analyzed by a multi-agent panel of physics specialists, yielding critique.

\subsection{PRISM: Producing the Dense Signal}
\label{sec:method:prism}

PRISM (Process-Reward Interpretation via Subsystem Mediation) is the densification machinery of MDForge: it converts the single terminal scalar $r^*_\pi$ into dense signals, 
\begin{equation}
r^*_\pi \;\xrightarrow{\;\text{PRISM}\;}\; \bigl(D_\pi,\, c_{\text{pre}},\, c_{\text{post}}\bigr),
\label{eq:densification}
\end{equation}
where $D_\pi$ is a $K$-tuple of per-stage physics diagnostics extracted from the simulator ($K{=}4$ canonical stages), and $c_{\text{pre}}, c_{\text{post}}$ are typed pre- and post-execution critiques aggregated from a panel of $J{=}3$ physics specialists. 

\paragraph{Per-stage physics diagnostics.}
The $K{=}4$ stages (Prep, Equilibration, Production sampling, Analysis) each run against a well-defined physical objective and expose interpretable diagnostics at their boundary. We attach to each stage a physics-grounded structured diagnostic: a typed record of canonical observables (e.g., force-field self-consistency at Prep, ergodicity and PME accuracy at Production, free-energy convergence at Analysis), extracted directly from the simulator's output rather than synthesized by an LLM. Concatenated across stages, these form $D_\pi$. Only Production phase incurs extensive GPU-hour cost.

\paragraph{Multi-agent debate over physics subsystems.}
$D_\pi$ is not yet actionable: a single measurement typically reflects several superimposed causes (force-field error, integrator instability, restraint misplacement, unconverged estimator) that no generic critic can disentangle. We delegate interpretation to the panel of $J{=}3$ specialist LLM agents, holding fixed, non-overlapping jurisdictions over canonical MD subsystems: \emph{Force Field}, \emph{Sampling}, and \emph{Analysis}. They deliberate in two rounds with cross-visibility~\citep{du2024debate}, and an aggregator $\mathcal{A}_\rho$ collapses their opinions into a single typed critique, weighted by per-expert reputations $\rho = (\rho_1, \ldots, \rho_J)$ so the panel can downweight historically-miscalibrated specialists rather than equal-averaging them with reliable ones; the update rule for $\rho$ is given in \S\ref{sec:method:writer}. Before the panel sees $\pi$, a tool-using Engineer agent debugs engineering faults (uncaught exceptions, missing files, mis-called APIs) in a sandbox without altering methodological choices, reserving panel deliberation for failures admitting physical attribution.

The panel is invoked at two points per trial: pre-execution it reviews $\pi$ and produces $c_{\text{pre}}$, a cheap screen the multi-agent system can act on before burning extensive running cost; post-execution it reviews the pipeline's execution results $B_\pi$ (predicted free energies with the corresponding accuracy and ranking metrics), producing
\begin{equation}
c_{\text{post}} \;=\; \mathcal{A}_\rho\!\bigl(\,\pi,\, B_\pi\,\bigr).
\label{eq:critique}
\end{equation}
Together with $D_\pi$, the pair $(c_{\text{pre}}, c_{\text{post}})$ completes the PRISM densification map of Equation \eqref{eq:densification}. The panel is advisory: only hard signals (Layer-1 rejection, divergence, timeout) gate execution.

\subsection{Code Agent Update and Reputation Loop}
\label{sec:method:writer}

\paragraph{Code agent update.}
Before the first trial, the panel holds a one-off design discussion over the task description $T$ (with web-search access), and its aggregated recommendations seed the Code agent's initial proposal. After each subsequent trial the Code agent regenerates the pipeline conditioned on the context bundle
\begin{equation}
\mathcal{C}_t \;=\; \bigl\{\, T,\; \pi_t,\; K_t,\; H_t\,\bigr\},
\label{eq:context}
\end{equation}
where $K_t = \{c_{l1,t},\, c_{\text{pre},t},\, c_{\text{post},t}\}$ is the critique set for trial $t$ (Layer-1 static check, pre-execution panel, post-execution panel on the benchmark) and $H_t$ is a compact trial history over all earlier trials (per-stage statuses and headline benchmark metrics). The per-stage diagnostic $D_{\pi_t}$ and terminal reward $r^*_{\pi_t}$ enter via the narrative text of $c_{\text{post},t}$ and the headline metrics in $H_t$; the reputation $\rho_t$ does not enter the Code agent's view directly and only shapes the aggregated critique through $\mathcal{A}_\rho$. The prior pipeline $\pi_t$ is included in $\mathcal{C}_t$ so revisions can remain localized edits when feasible rather than wholesale rewrites; the full edit protocol and prompt template are in Appendix~\ref{app:method-details}.

\paragraph{Reputation loop.}
A slow per-task loop maintains $\rho_t$ by per-expert agreement between $c_{\text{pre},t}$ and $c_{\text{post},t}$: experts whose pre-trial calls are validated by post-execution evidence accumulate weight within the task. The update rule and its in-task convergence are in Appendix~\ref{app:method-details}. 
\input{tables/main_results}

%% file: tables/main_results.tex
\providecommand{\runyes}{\textcolor{green!55!black}{\ensuremath{\checkmark}}}
\providecommand{\runno}{\textcolor{red!65!black}{\ensuremath{\times}}}
\newsavebox{\runpartbox}
\sbox{\runpartbox}{\textcolor{orange!85!black}{%
  \tikz[baseline=-0.55ex]{%
    \draw[line width=0.5pt] (0,0) circle (0.55ex);%
    \fill (0,0) -- (0,0.55ex) arc (90:-90:0.55ex) -- cycle;%
  }%
}}
\providecommand{\runpart}{\usebox{\runpartbox}}

\begin{table*}[!t]
  \centering
  \small
  \setlength{\tabcolsep}{3pt}
  \renewcommand{\arraystretch}{1.10}
  \begin{tabular*}{0.95\textwidth}{@{\extracolsep{\fill}}llccccccc@{}}
    \toprule
    \multirow{2}{*}{Host} & \multirow{2}{*}{Method} & \multirow{2}{*}{Runnable} & \multicolumn{3}{c}{Train (4 guests)} & \multicolumn{3}{c}{Test (held-out guests)} \\
    \cmidrule(lr){4-6} \cmidrule(lr){7-9}
    & & & $R^2$ & Spearman $\rho$ & Kendall $\tau$ & $R^2$ & Spearman $\rho$ & Kendall $\tau$ \\
    \midrule
    \multirow{7}{*}{CB[7]}
      & No feedback                                                  & \runno   & -- & -- & -- & -- & -- & -- \\
      & LLM critic~\citep{madaan2023selfrefine}                      & \runno   & -- & -- & -- & -- & -- & -- \\
      & Step-level feedback~\citep{yao2023react}                     & \runpart & 0.16 & 0.40 & 0.33 & 0.14 & 0.21 & 0.16 \\
      & Trial-level feedback~\citep{shinn2023reflexion}              & \runyes  & 0.29 & 0.40 & 0.33 & 0.32 & 0.32 & 0.24 \\
    \cmidrule{2-9}
      & \textbf{MDForge}                                             & \runyes  & \textbf{0.68} & \textbf{0.80} & \textbf{0.67} & \textbf{0.58} & \textbf{0.68} & \textbf{0.56} \\
      & \quad w/o Multi-expert debate                                & \runyes  & 0.73 & 0.80 & 0.67 & 0.53 & 0.33 & 0.33 \\
      & \quad w/o Stage diagnostics                                  & \runyes  & 0.34 & 0.40 & 0.33 & 0.45 & 0.10 & 0.16 \\
    \midrule
    \multirow{7}{*}{OAH}
      & No feedback                                                  & \runno   & -- & -- & -- & -- & -- & -- \\
      & LLM critic~\citep{madaan2023selfrefine}                      & \runno   & -- & -- & -- & -- & -- & -- \\
      & Step-level feedback~\citep{yao2023react}                     & \runpart & 0.23 & 0.40 & 0.33 & 0.13 & 0.20 & 0.20 \\
      & Trial-level feedback~\citep{shinn2023reflexion}              & \runyes  & 0.47 & 0.60 & 0.33 & 0.21 & 0.20 & 0.20 \\
    \cmidrule{2-9}
      & \textbf{MDForge}                                             & \runyes  & \textbf{0.99} & \textbf{1.00} & \textbf{1.00} & \textbf{0.43} & \textbf{0.30} & \textbf{0.20} \\
      & \quad w/o Multi-expert debate                                & \runyes  & 0.59 & 0.40 & 0.33 & 0.26 & 0.10 & 0.00 \\
      & \quad w/o Stage diagnostics                                  & \runyes  & 0.12 & 0.40 & 0.33 & 0.10 & $-$0.10 & 0.00 \\
    \midrule
    \multirow{7}{*}{CBClip}
      & No feedback                                                  & \runno   & -- & -- & -- & -- & -- & -- \\
      & LLM critic~\citep{madaan2023selfrefine}                      & \runno   & -- & -- & -- & -- & -- & -- \\
      & Step-level feedback~\citep{yao2023react}                     & \runpart & 0.12 & 0.40 & 0.33 & 0.03 & 0.09 & 0.07 \\
      & Trial-level feedback~\citep{shinn2023reflexion}              & \runyes  & 0.54 & 0.40 & 0.33 & 0.09 & 0.26 & 0.20 \\
    \cmidrule{2-9}
      & \textbf{MDForge}                                             & \runyes  & \textbf{0.63} & \textbf{0.80} & \textbf{0.67} & \textbf{0.54} & \textbf{0.60} & \textbf{0.47} \\
      & \quad w/o Multi-expert debate                                & \runyes  & 0.71 & 0.80 & 0.67 & 0.23 & 0.37 & 0.33 \\
      & \quad w/o Stage diagnostics                                  & \runyes  & 0.54 & 0.40 & 0.33 & 0.03 & 0.31 & 0.20 \\
    \bottomrule
  \end{tabular*}
  \caption{\textbf{Results on SAMPL host–guest binding benchmarks}: CB[7] ($n_\mathrm{held}{=}10$), OAH ($n_\mathrm{held}{=}5$), CBClip ($n_\mathrm{held}{=}6$), with 4 training guests selected at experimental-$\Delta G$ quintile positions. We report $R^2$, Spearman $\rho$, and Kendall $\tau$ against experimental $\Delta G$, for the best of $N{=}5$ successful trials per host (selected by training-set $\tau$). \emph{Runnable} summarizes how often a method produces an executable pipeline across the $N{=}5$ trials: \runyes{} = all~5, \runpart{} = 1--2, \runno{} = none. ``--'' marks methods with no runnable trial on any host.}
  \vspace{-10pt}
  \label{tab:main}
\end{table*}

%% file: sections/experiments.tex
\section{Experiments}
\label{sec:experiments}

\input{tables/per_trial}

\subsection{Setup}
\label{sec:exp:setup}

\paragraph{Task.}
We test MDForge on the SAMPL host–guest binding free-energy challenges~\citep{sampl4cb7}, a widely used MD benchmark and the standard tractable proxy for the protein–ligand binding problem that drives drug discovery. A rigid macrocyclic host plays the role of the protein pocket, and the task is to compute the binding free-energy $\Delta\hat{G}$ of a small molecule guest against a known experimental reference. The host–guest setting preserves the thermodynamic machinery of protein–ligand binding (water displacement, ion solvation, anharmonic guest sampling) while removing protein flexibility, so accuracy here is a necessary precondition for the full-protein setting and a fair stress test of an MD design agent. We evaluate on three hosts of distinct chemistry: \emph{CB[7]} (SAMPL4), \emph{OAH} (SAMPL4), and \emph{CBClip} (SAMPL5). For each host the guests are split into a 4-guest training set (visible to the verbal RL feedback) and the remainder as a held-out test set. The docked pose for each guest is supplied; the experimental reference $\Delta G_{\exp}$ is held aside from the agent. We use a cheap-MD configuration: each guest evaluation runs in $\approx 2$ GPU-hours on a single A40 node. 

\paragraph{Methods.}
Baselines form a capability ladder organized by the type of feedback available during pipeline construction: (1) \emph{No feedback}: one-pass code generation with no critique and no execution signal; (2) \emph{LLM critic}~\citep{madaan2023selfrefine}: an auxiliary LLM reviews and rewrites the draft, but no code is executed; (3) \emph{Step-level feedback}~\citep{yao2023react}: tool calls return intermediate results during reasoning, enabling partial in-trial recovery but no memory across trials; (4) \emph{Trial-level feedback}~\citep{shinn2023reflexion}: a natural-language summary of each completed trial's outcome conditions the next trial; and (5) \emph{MDForge}: trial-level feedback augmented with PRISM (stage diagnostics and multi-expert debate). Each method runs until $N{=}5$ successful trials accumulate per host.

\input{tables/comp_tool_call}

\begin{figure*}[!t]
  \centering
  \includegraphics[width=\linewidth]{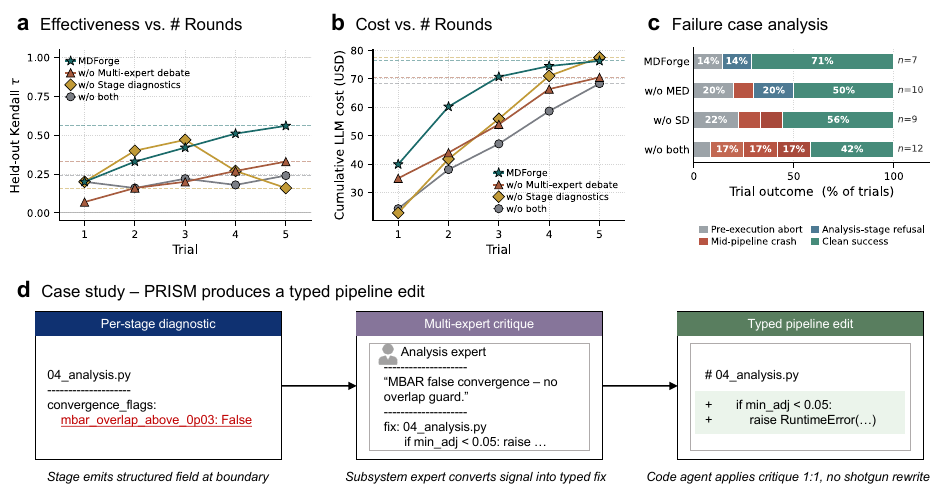}
  \caption{\textbf{PRISM mechanism on CB[7].} (a) Per-trial held-out Kendall $\tau$ over $N{=}5$ trials (CB[7] row of Table~\ref{tab:main}). MDForge improves monotonically to $\tau{=}0.56$; w/o Stage diagnostics peaks at trial 3 then collapses to $0.16$, overshooting without a typed signal. (b) \crpol{Cumulative \emph{LLM token} spend ranges only $\$68$ to $\$78$ across methods, so panel (a)'s gain carries no token-cost premium.} (c) PRISM types the failures: mid-pipeline crashes shrink as components are added, while analysis-stage refusal ($51\% \to 0\%$) appears only with stage diagnostics, since the convergence guard creates that category. (d) One typed signal yields one localized edit, not a shotgun rewrite: the Analysis stage emits \texttt{mbar\_overlap\_above\_0p03 = False}, the analysis expert proposes a guard, and the Code agent applies a one-line fix at the named location.}
  \label{fig:diagnosis}
\end{figure*}

\input{tables/tool_comparison}

\subsection{Main Results}
\label{sec:exp:main}

\paragraph{Overview Results.}
Table~\ref{tab:main} separates two questions: \emph{can the agent produce a runnable MD pipeline}, and \emph{how much ranking signal does it then recover}? The first already filters most of the ladder: No-feedback and LLM-critic baselines fail at coding on every host (0/5), Step-level feedback succeeds only intermittently (1 to 2 of 5), and reliable code emerges only with cross-trial memory (Trial-level feedback and MDForge, 5/5 everywhere).
Among methods that do code, MDForge attains a held-out Kendall $\tau$ of $0.56$ on CB[7] and $0.47$ on CBClip against $0.24$ and $0.20$ for the Trial-level baseline, \emph{more than doubling the ranking signal that transfers from the 4-guest training set to held-out guests}. On CBClip this places MDForge in the performance band of the SAMPL5 BEDAM and SOMD human submissions~\citep{sampl5cbclip}. OAH is an information-limited exception ($n_{\mathrm{held}}{=}5$, narrow $\Delta G$ window): all coding methods cluster at $\tau \approx 0.20$. We use rank-based metrics because MD predictions carry method-specific force-field offsets (e.g., GAFF over-binds cationic CB[7] guests).

\paragraph{Per-Trial Results.}

Table~\ref{tab:per-trial} reports held-out Kendall $\tau$ for each of the $N{=}5$ successful trials per host, rather than the selected trial alone. The separation between MDForge and the baselines is a property of the whole distribution, not of the selection rule: on every host the MDForge trial-mean is roughly double the best baseline trial-mean ($0.40$ vs.\ $0.20$ on CB[7], $0.28$ vs.\ $0.12$ on OAH, $0.41$ vs.\ $0.20$ on CBClip), and the gap between the two means is about $1.4\times$ the MDForge standard deviation. A typical, unselected MDForge trial therefore outperforms a typical baseline trial by more than MDForge's own run-to-run noise. Selection accounts for the distance between the trial-mean and the reported endpoint (CB[7]: $0.40 \to 0.56$), but not for the margin over the baselines, which is already present at the mean.

\paragraph{Comparison Against the Fixed-Toolbox Paradigm.}

Prior MD agents such as MDCrow~\citep{mdcrow2024} and MDAgent~\citep{ma2026mdagent} orchestrate a predefined toolbox rather than emitting open-ended code, and are not directly runnable on host--guest binding free-energy pipelines. We therefore reproduce their \emph{paradigm} rather than their systems: a \texttt{tool-calling} baseline exposing $15$ MD tools that cover the modules a toolbox would plausibly provide (force-field and charge assignment, solvation, minimisation, NVT/NPT equilibration, APR restraints, $\lambda$-schedule construction, production, and MBAR/BAR-TI estimation), driven by the same backbone, the same simulator interface, and the same budget as MDForge. Open-ended code generation is the only variable that changes. We report two variants: one in which the agent only selects and orders tools, leaving their internal parameters at defaults, and one in which it may additionally tune the parameters of tools.

Table~\ref{tab:toolcalling} shows that code generation dominates the toolbox on CB[7] ($0.56$ vs.\ $0.22$) and on OAH, where both toolbox variants degenerate to $\tau{=}0.00$. On CBClip a well-tuned toolbox matches MDForge, but only reaches $0.47$ through hyperparameter tuning layered on the fixed-parameter baseline ($0.33 \to 0.47$), whereas MDForge attains it directly. The gap is exactly the set of design choices a hand-enumerated toolbox cannot express.

\subsection{Diagnosis}
\label{sec:exp:diagnosis}

Figure~\ref{fig:diagnosis} asks whether MDForge's CB[7] endpoint comes from the claimed mechanism: that verbal RL can turn PRISM's typed signals into localized pipeline edits.

\paragraph{Effectiveness (figure~\ref{fig:diagnosis}a).}
MDForge climbs monotonically to $\tau{=}0.56$, while the debate-only ablation peaks at $0.47$ in trial 3 then collapses to $0.16$. Without per-stage signal, the agent cannot tell a good edit from a regression and discards a working pipeline. The other two methods stall near $\tau \approx 0.20$. PRISM's value is \emph{keeping} a high $\tau$.

\paragraph{Cost (figure~\ref{fig:diagnosis}b).}
All four methods land within $\$68$ to $\$78$ at trial 5. MDForge's early per-trial token overhead is amortized once edits become localized rather than wholesale rewrites, so panel-(a)'s ranking gain carries no cost premium.

\paragraph{Failure typing (figure~\ref{fig:diagnosis}c).}
PRISM \emph{types} the failures rather than lowering their count. Mid-pipeline crashes shrink from $51\%$ w/o both to $0\%$ on MDForge, while analysis-stage refusal appears only with stage diagnostics: it is the convergence guard explicitly declining to emit a silent MBAR false-convergence. The $56\%$ clean-success on w/o Stage diagnostics therefore includes outcomes the guard would have refused.

\paragraph{Case study (figure~\ref{fig:diagnosis}d).}
One trial follows the chain end to end: the Analysis stage emits \texttt{mbar\_overlap\_above\_0p03 = False}; the analysis specialist proposes a guard; the Code agent applies a one-line edit at the named location, not a shotgun rewrite. Panel~(c)'s blue segments aggregate this mechanism across trials, making panel~(a)'s gain reproducible rather than lucky.

\begin{figure*}[!t]
  \centering
  \includegraphics[width=\linewidth]{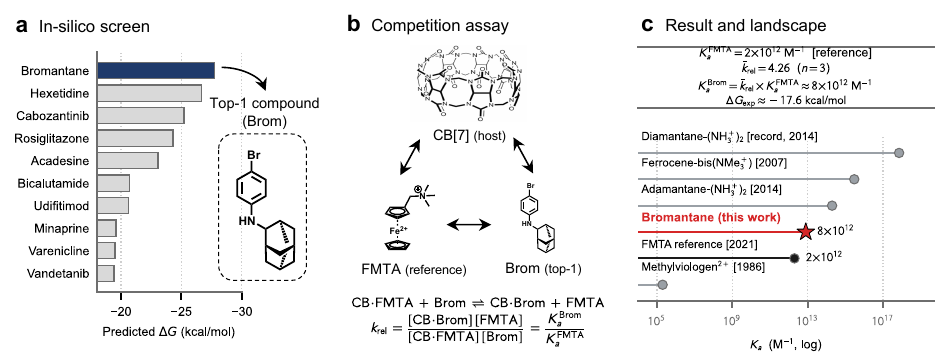}
  \caption{\textbf{End-to-end discovery from in-silico screening to wet-lab confirmation.} (a) Ten unseen candidate guests ranked by MDForge-predicted binding free-energy; top-1 is Bromantane (Brom). (b) Competition $^{1}$H NMR assay: CB[7], the picomolar reference guest ferrocenylmethyl-trimethylammonium (FMTA; $K_a^{\mathrm{FMTA}} \approx 2 \times 10^{12}$~M$^{-1}$~\citep{alnajjar2021cb7scale}), and Brom co-equilibrate. (c) Top: measured $\bar{k}_{\mathrm{rel}} = 4.26$ ($n{=}3$) yields $K_a^{\mathrm{Brom}} \approx 8 \times 10^{12}$~M$^{-1}$ ($\Delta G_{\mathrm{exp}} \approx -17.6$~kcal/mol). Bottom: Brom (red star) plotted against published CB[7] binders spanning $K_a$ from $10^5$ to $10^{17}$~M$^{-1}$~\citep{cao2014diamantane,rekharsky2007ferrocene,moghaddam2011bco,liu2005adamantane,mock1986viologen}.}
  \label{fig:wetlab}
\end{figure*}

\subsection{Does MDForge Build Like a Human Expert?}
\label{sec:exp:expert}

Beyond agent-vs-agent comparison, the chemistry-credibility check is whether MDForge's \emph{pipeline} is the kind of pipeline a human MD expert would actually design. As our expert reference we use the canonical \texttt{pAPRika} APR pipeline from the Gilson lab~\citep{henriksen2015cb7,slochower2019paprika}, the open-source reference implementation that has served as the \textit{de facto} standard for host–guest free energy calculations on cucurbit[n]uril and related systems for nearly a decade. Table~\ref{tab:tool-comparison} contrasts per-stage tool selection on SAMPL4 CB[7] across three pipelines: this expert reference; an AI + non-expert pipeline that a chemistry non-expert assembled with LLM coding assistance; and MDForge running autonomously. The reading splits the two AI-touched columns in opposite directions: MDForge \emph{stays in the expert's methodological family} (GAFF2 + AM1-BCC, APR umbrella, MBAR), departing only on reliability-flavored engineering (italicized in the table). The AI + non-expert pipeline, in contrast, \emph{diverges to an alternative method family} (z-PMF umbrella rather than APR with dummy-atom geometry), a defensible but methodologically distinct route that LLM coding assistance plausibly leads a non-expert toward.

The \emph{Performance} row sharpens the comparison along the same gradient. MDForge attains $\rho{=}0.68$, $\tau{=}0.56$, $R^2{=}0.58$ on the SAMPL4 CB[7] guests, \emph{recovering 78--82\% of the expert's ranking utility} ($\rho{=}0.83$, $\tau{=}0.68$, $R^2{=}0.74$) without a human in the loop, and \emph{beating the AI~+~non-expert pipeline on all three rank-correlation metrics} ($\rho{=}0.61$, $\tau{=}0.47$, $R^2{=}0.44$). Absolute-error metrics are dominated by force-field-specific systematic biases (e.g., the well-known GAFF CB[7]–cation over-binding) and therefore do not present a truly "apples-to-apples" cross-pipeline comparison.

\subsection{Prospective Wet-Lab Validation}
\label{sec:exp:wetlab}

Retrospective benchmark accuracy is necessary but not sufficient: a useful pipeline must also \emph{deploy} prospectively. We applied the best MDForge CB[7] pipeline to ten unseen candidate guests drawn from the compound bank we extracted from ChEMBL~\citep{zdrazil2024chembl} and DrugBank~\citep{knox2024drugbank}, ranked them by predicted $\Delta\hat{G}$, and submitted the top-1 (Bromantane, "Brom"; Figure~\ref{fig:wetlab}a) for wet-lab measurement. Because $K_a$ in the picomolar regime exceeds what direct isothermal titration calorimetry can resolve, we used competition $^{1}$H NMR against a reference guest of known affinity (Figure~\ref{fig:wetlab}b). Co-equilibrating Brom with CB[7] and the canonical picomolar reference ferrocenylmethyl-trimethylammonium (FMTA; $K_a^{\mathrm{FMTA}} \approx 2 \times 10^{12}$~M$^{-1}$~\citep{alnajjar2021cb7scale}) establishes the exchange $\mathrm{CB}\!\cdot\!\mathrm{FMTA} + \mathrm{Brom} \rightleftharpoons \mathrm{CB}\!\cdot\!\mathrm{Brom} + \mathrm{FMTA}$, whose relative constant $K_{\mathrm{rel}} = K_a^{\mathrm{Brom}} / K_a^{\mathrm{FMTA}}$ is read directly from the bound and unbound NMR integrations of both guests (the free-host concentration cancels). Averaging across three independent guest-ratio mixtures yields $K_{\mathrm{rel}} = 4.26$ ($n{=}3$ samples run at different molar ratios), hence $K_a^{\mathrm{Brom}} \approx 8 \times 10^{12}$~M$^{-1}$ ($\Delta G_{\mathrm{exp}} \approx -17.6$~kcal/mol; Figure~\ref{fig:wetlab}c, top), approximately four-fold tighter than the FMTA reference. The landscape (Figure~\ref{fig:wetlab}c, bottom) places Brom in the picomolar high-affinity tier of published CB[7] binders, comparable to deliberately-engineered ferrocene and adamantane di-ammonium guests, while remaining $\sim$5 orders of magnitude below the current record holder (diamantane-bis(ammonium))~\citep{cao2014diamantane}; all entries above Brom on this scale were obtained through years of human-driven design.

We tested only the top-1 candidate, not the other nine: this single wet-lab measurement is intended to demonstrate that MDForge can translate in-silico design into a real prospective scientific discovery, rather than to make the discovered molecule itself the primary contribution of this work. Furthermore, no claims are made regarding the expected binding affinity or validation of the full rankings across all ten hits identified from in-silico screening.

%% file: tables/per_trial.tex
\begin{table}[t]
  \centering
  \small
  \setlength{\tabcolsep}{3pt}
  \renewcommand{\arraystretch}{1.10}
  \begin{tabular*}{\columnwidth}{@{\extracolsep{\fill}}llccccc@{}}
    \toprule
    \multirow{2}{*}{Host} & \multirow{2}{*}{Method} & \multicolumn{5}{c}{Trial} \\
    \cmidrule(lr){3-7}
    & & 1 & 2 & 3 & 4 & 5 \\
    \midrule
    \multirow{3}{*}{CB[7]}
      & Step-level  & --   & --   & --   & 0.16 & 0.16 \\
      & Trial-level & 0.20 & 0.16 & 0.22 & 0.18 & 0.24 \\
      & \textbf{MDForge} & 0.20 & 0.33 & 0.42 & 0.51 & 0.56 \\
    \midrule
    \multirow{3}{*}{OAH}
      & Step-level  & --   & --   & 0.20 & --   & 0.00 \\
      & Trial-level & 0.00 & 0.20 & 0.20 & 0.20 & 0.00 \\
      & \textbf{MDForge} & 0.40 & 0.40 & 0.20 & 0.20 & 0.20 \\
    \midrule
    \multirow{3}{*}{CBClip}
      & Step-level  & --   & --   & --   & 0.07 & 0.07 \\
      & Trial-level & 0.07 & 0.20 & 0.33 & 0.20 & 0.20 \\
      & \textbf{MDForge} & 0.20 & 0.33 & 0.60 & 0.47 & 0.47 \\
    \bottomrule
  \end{tabular*}
  \caption{\textbf{Per-trial held-out Kendall $\tau$ across all $N{=}5$ successful trials.} ``--'' marks a trial that did not yield a runnable pipeline. }
  \label{tab:per-trial}
\end{table}

%% file: tables/comp_tool_call.tex
\begin{table}[t]
  \centering
  \small
  \setlength{\tabcolsep}{3pt}
  \renewcommand{\arraystretch}{1.10}
  \begin{tabular*}{\columnwidth}{@{\extracolsep{\fill}}lccc@{}}
    \toprule
    Method & CB[7] & OAH & CBClip \\
    \midrule
    Tool-calling, fixed hyperparameters & 0.22 & 0.00 & 0.33 \\
    Tool-calling, tuned hyperparameters & 0.28 & 0.00 & 0.47 \\
    \textbf{MDForge} (open-ended code gen.) & \textbf{0.56} & \textbf{0.20} & \textbf{0.47} \\
    \bottomrule
  \end{tabular*}
  \caption{\textbf{Open-ended code generation vs.\ the fixed-toolbox paradigm.} Held-out Kendall $\tau$. The tool-calling rows reproduce the paradigm of prior MD agents~\citep{mdcrow2024,ma2026mdagent} under matched backbone, simulator, and budget; they are not runs of those systems.}
  \label{tab:toolcalling}
\end{table}

%% file: tables/tool_comparison.tex
\providecommand{\dvg}[1]{\textcolor{blue!55!black}{\textbf{\emph{#1}}}}
\providecommand{\perfval}[1]{\textbf{#1}}

\begin{table*}[t]
  \centering
  \scriptsize
  \setlength{\tabcolsep}{4pt}
  \renewcommand{\arraystretch}{1.18}
  \begin{tabular}{@{}p{0.10\linewidth}p{0.27\linewidth}p{0.27\linewidth}p{0.27\linewidth}@{}}
    \toprule
    & \multicolumn{3}{c}{\textcolor{gray!60}{\rule[0.55ex]{3em}{0.4pt}}\;\textsf{\itshape decreasing human involvement}\;\textcolor{gray!60}{$\xrightarrow{\hspace*{3em}}$}} \\
    \cmidrule(lr){2-4}
    Stage
      & \textbf{Pure expert} \newline {\tiny canonical \texttt{pAPRika} APR pipeline}
      & \textbf{AI + non-expert} \newline {\tiny non-expert + LLM coding assistance}
      & \textbf{Pure AI} \newline {\tiny MDForge, autonomous} \\
    \midrule

    Preparation
      & GAFF2 + AM1-BCC; pre-charged CB[7] mol2; TIP3P + Joung-Cheatham ions; $\geq$12~\AA{} pad; \texttt{taproom} dummy-atom APR geometry.
      & \dvg{Glide XP docking}; \texttt{antechamber} + AM1-BCC; GAFF/GAFF2/\dvg{OpenFF~2.2}; TIP3P or \dvg{OPC}; 12~\AA{} pad; \dvg{no dummy geometry}.
      & Same GAFF2 + AM1-BCC, with \dvg{\texttt{nz/n3}$\to$\texttt{n4} widening fix}; \texttt{tleap} 18~\AA{} pad; \dvg{auto-scans \texttt{tleap.log}}. \\ 

    Equilibrate
      & NVT 10$\to$298~K (250~ps, restrained) + NPT 1~atm ramp-down (500~ps); MC barostat; HMR/SHAKE, 2--4~fs; PME.
      & Minimise; \dvg{5-stage heat 50$\to$298~K (5$\times$10~ps)}; \dvg{0.5~ns NPT}; optional \dvg{``staged''} fallback for difficult poses.
      & NVT heat + NPT density equilibration; HMR via OpenMM \texttt{hydrogenMass}. \\

    Production
      & APR umbrella: 6 attach + 14--18 pull to $r_{\max}\!\approx\!18$~\AA; analytical release with SSC; 5--10~ns/win; HMR + 4~fs.
      & \dvg{5~ns unbiased NVT}, then \dvg{z-PMF umbrella (not APR)}: \dvg{24 windows} $z\!\in\![-2, +20]$~\AA{} at 1~\AA; \dvg{xy cylinder restraint} ($r\!=\!2$~\AA); 0.1+0.3~ns/win.
      & APR umbrella, \dvg{budget-tuned}: 7 attach ($\lambda$ schedule) + 14 pull at 0.5~\AA; $r_{\max}\!=\!r_{\mathrm{bound}}\!+\!7$~\AA; \dvg{50/10~ps equil floor}; per-win velocity reseed. \\

    Analysis
      & MBAR~\citep{shirts2008mbar} on 22 windows; auto-correlation subsampling; SSC + symmetry; block-convergence flag ($<$0.5~kcal/mol).
      & \texttt{pymbar 4.0} PMF; \dvg{Woo-Roux SSC} from measured xy area; reports raw + corrected $\Delta G$; \dvg{no MBAR overlap / convergence guards}.
      & MBAR, \dvg{stricter guards}: full $K\!\times\!K$ overlap; block thresh $\max(2\sigma, 0.5)$~kcal/mol; \dvg{stat$+$sys uncertainty}; \dvg{p$K_a$-aware protonation correction}. \\

    \specialrule{0.8pt}{2pt}{2pt}
    Performance
      & $\rho{=}\perfval{0.83}$, $\tau{=}\perfval{0.68}$, $R^2{=}\perfval{0.74}$ 
      & $\rho{=}\perfval{0.61}$, $\tau{=}\perfval{0.47}$, $R^2{=}\perfval{0.44}$ 
      & $\rho{=}\perfval{0.68}$, $\tau{=}\perfval{0.56}$, $R^2{=}\perfval{0.58}$ \\

    \bottomrule
  \end{tabular}
  \caption{\textbf{Per-stage tool selection.} Tool selection on SAMPL4 CB[7] across (i) \emph{pure expert}, the canonical \texttt{pAPRika} APR pipeline~\citep{slochower2019paprika}; (ii) \emph{AI + non-expert}, a chemistry non-expert's pipeline assembled with LLM coding assistance; (iii) \emph{pure AI}, MDForge autonomous. \dvg{Navy bold italics} mark choices that depart from the expert default. MDForge stays in the expert's family (GAFF2/AM1-BCC, APR, MBAR) with only reliability-flavored engineering deviations; the AI + non-expert pipeline diverges (z-PMF umbrella, no MBAR guards). Performance reports rank metrics ($\rho$, $\tau$, $R^2$), which wash out force-field-specific systematic biases on absolute $\Delta G$.}
  \label{tab:tool-comparison}
\end{table*}

%% file: sections/conclusion.tex
\section{Conclusion}
\label{sec:conclusion}

We introduced MDForge, an LLM agent that designs molecular dynamics pipelines as open-ended code under verbal RL. Its sparse terminal reward is densified by PRISM into per-stage diagnostics and a typed, subsystem-attributable expert critique. On SAMPL host–guest benchmarks, MDForge dominates other LLM-agent designs in accuracy. Its per-stage tool choices track those of the human-expert submissions. Deployed prospectively on an unseen compound library, MDForge discovered a novel CB[7] binder. Wet-lab competition NMR against the FMTA picomolar reference measures $K_a \approx 8 \times 10^{12}$~M$^{-1}$, placing it in the high-affinity (picomolar) regime.

\section*{Acknowledgments}
This work was partially supported by the NSF under grants IIS-2533550, IIS-2321504, IIS-2217239, CNS-2426514, and CMMI-2146076, Notre Dame Strategic Framework Research Grant (2025), and Notre Dame Poverty Research Package (2025). Any expressed opinions, findings, and conclusions or recommendations are those of the authors and do not necessarily reflect the views of the sponsors.

%% file: sections/limitations.tex
\section*{Limitations}
\label{sec:limitations}

\paragraph{Benchmark scope.}
We benchmark MDForge on three host–guest systems from the SAMPL series (CB[7], OAH, CBClip), a restricted slice of the broader MD design space. The temporal-staging and subsystem-expert decompositions are in principle agnostic to the target system class. But the absence of equally mature open benchmarks for protein–ligand affinity, membrane protein insertion, and other binding regimes bounds our empirical reach. Future work can apply the same framework to broader binding benchmarks (e.g., FEP+, PDBbind) and to non-binding MD applications such as conformational free energy surfaces.

\paragraph{MD fidelity.}
We operate in a cheap-MD configuration ($\approx$2~GPU-hours per guest) to keep the per-task budget tractable for systematic evaluation. The framework is in principle compatible with longer production sampling, higher-accuracy force fields, and explicit polarization. We expect the comparative method ranking in Table~\ref{tab:main} to persist across configurations, since the limiting factor in our regime is the verbal RL update rather than the underlying MD fidelity. Verifying this at higher fidelity is left to future work.

\paragraph{Scope of the wet-lab demonstration.}
The prospective wet-lab measurement (\S\ref{sec:exp:wetlab}) covers exactly one data point at the top of the predicted ranking. We do not validate the full ten-compound ranking individually, nor confirm binding affinity for any of the other predicted binders. Those measurements are out of scope here. What we do validate is that the framework retrieves a real high-affinity CB[7] binder at the top of an unseen library, end-to-end. We treat this as the right scope for a first prospective demonstration; broader prospective screens against multiple hosts are deferred to follow-up work.

%% file: sections/ethics.tex
\section*{Ethical Considerations}
\label{sec:ethics}

\paragraph{Dual-use risk.}
MDForge automates the design of MD pipelines for binding affinity prediction. The same capability that accelerates therapeutic discovery could in principle be repurposed to design harmful binders. Our prospective wet-lab result sharpens this concern from a theoretical possibility to an operational one. The framework's direct output is a simulation protocol (Python code), not a molecule. The locus of dual-use control therefore sits upstream (in the candidate library and the choice of target) and downstream (in the interpretation and deployment), not at MDForge itself. Responsible deployment requires governance over those layers, consistent with established norms in computational chemistry and structure-based drug discovery~\citep{boiko2023coscientist,bran2024chemcrow}.

\paragraph{Scope of the wet-lab demonstration.}
The compound bank from which the top-1 candidate was drawn was constructed in separate work and is not a curated set of pharmacologically active species. CB[7] is a synthetic macrocyclic host, not a biological target. The discovered binder therefore has no direct therapeutic use, and the experiment should not be read as a drug-discovery claim. The bank composition is governed by the originating project; the curated compound bank will be released in a follow-on publication.

\paragraph{Computational footprint.}
End-to-end MD-based virtual screening is energy-intensive. We deliberately operate in a cheap-MD configuration ($\approx$2~GPU-hours per guest) and a small trial budget, both of which lower the per-discovery energy cost relative to traditional expert-design iteration. Useful community directions for further reducing this footprint include energy-aware scheduling, surrogate filters that defer expensive sampling, and shared baselines that quantify energy-per-prediction relative to human-designed pipelines.

\paragraph{Reproducibility and auditability.}
AI-designed MD pipelines must be auditable by domain experts before being treated as scientific instruments. MDForge produces Python code as its action rather than opaque numerical parameters, so individual pipelines are inspectable. The generated code, however, is often long and stylistically idiosyncratic, which raises the audit burden. We release the agent code, per-trial pipeline artifacts, multi-expert prompts, and the full LLM configuration. Promising future directions include pipeline-summarization tools that compress agent-generated code into expert-readable protocol descriptions, and automated provenance tracking that links each design choice back to the typed critique that motivated it.

%% file: appendix/related_work.tex
\section{Comprehensive Related Work}
\label{app:related-extended}

\subsection{Molecular Dynamics}
\label{app:related:md}
MDForge sits atop the established methodological stack of physics-based MD rather than competing with any of its parts. The binding free-energy task we target is computed by the alchemical FEP/TI family with BAR/MBAR estimators~\citep{bennett1976bar,shirts2008mbar,mey2020bestpractices}; absolute free-energy workflows additionally rely on standard-state restraint corrections~\citep{boresch2003standardstate,gilson1997statistical} and have been benchmarked extensively in the relative regime~\citep{wang2015rbfe,cournia2017relative}. The executable pipelines our agent emits target mature engines (OpenMM, GROMACS, AMBER)~\citep{eastman2017openmm,abraham2015gromacs,case2023amber} and biomolecular force-field families (AMBER, GAFF/OpenFF, TIP3P/OPC water), which form the action vocabulary the LLM composes into rather than targets it tries to improve. A parallel line of work proposes neural surrogates for parts of the pipeline: ML force fields trained to replace classical potentials~\citep{behler2021nnp,unke2021mlff}, end-to-end structure predictors~\citep{jumper2021alphafold,wu2026proteor1}, and neural samplers that draw equilibrium configurations directly~\citep{noe2019boltzmann}. Each replaces a single slice (a force, a static structure, an equilibrium sample), but none yields the staged, diagnostic-emitting trajectory whose design MDForge automates. Earlier non-LLM frameworks (BioSimSpace, OpenFE, perses)~\citep{hedges2019biosimspace} also templatize parts of this workflow; MDForge instead emits the workflow itself as code, so the design space is the program-synthesis space rather than the parameter space of a fixed template.

\subsection{Autonomous Science Agents}
\label{app:related:agents}
LLM-driven scientific agents integrate literature search, hypothesis proposal, code synthesis, and outer-loop evaluation into runnable discovery pipelines~\citep{boiko2023coscientist,romera2024funsearch,sakana2024aiscientist,bran2024chemcrow,baek2024researchagent,schmidgall2025agentlab,chen2026longhorizon,ma2026policy4ood}, with parallel instantiations in materials discovery and autonomous laboratories~\citep{merchant2023gnome,szymanski2023alab}, biological experiment and protocol planning~\citep{qu2024crisprgpt,odonoghue2023bioplanner}, and software- or ML-engineering pipeline automation~\citep{yang2024sweagent,trirat2024automlagent,chan2024mlebench}. Closest to our setting, MD-specific agents have converged on a \emph{tool-calling} pattern in which an LLM orchestrates a fixed library of simulation engines, force fields, and analysis libraries into an executable workflow. MDCrow~\citep{mdcrow2024} provides a general-purpose toolset over force-field setup, simulation, and trajectory analysis, exposed to an LLM through LangChain-style tool calls. MDAgent~\citep{ma2026mdagent} extends the pattern with a case-based skill-and-memory module that retrieves prior task knowledge across trials, the most ambitious recent attempt at inter-trial improvement within the tool-calling paradigm. Similar tool-calling patterns extend to binding free-energy workflows~\citep{dynamate2025}, polymer and topological MD~\citep{polyjarvis2026,ding2025topolyagent}, NAMD-based protein simulations~\citep{chandrasekhar2025namdagent}, alloy design~\citep{ghafarollahi2025atomagents}, broader atomistic settings~\citep{vriza2026multiagentic,yang2026quasar}, and to fine-tuned LLM variants that emit MD scripts directly~\citep{shi2025mdagent,shi2026mdagent2}.
In contrast, MDForge treats MD pipeline design as \emph{open-ended code generation}, placing it in the lineage of program-synthesis agents that emit arbitrary executable code rather than select from a fixed tool library~\citep{wang2023voyager,romera2024funsearch}; unlike these, MDForge operates in a regime whose only supervisory signal is sparse (one terminal reward per full pipeline run) and expensive (GPU-hours of MD execution per trial).

\subsection{Verbal Reinforcement Learning}
\label{app:related:verbalrl}
In-context learning enables an LLM agent to reshape its behavior online by manipulating context rather than parameters, requiring no gradient update. Reflexion~\citep{shinn2023reflexion} formalized this as \emph{verbal reinforcement learning}: an LLM agent attempts a trial, receives a textual outcome label, generates a natural-language reflection, and consumes that reflection as additional context on the next trial; recent work further formalizes this verbal-feedback channel without scalar rewards~\citep{luo2025fcp}. Subsequent work extends this paradigm with reasoning-action interleaving~\citep{yao2023react,madaan2023selfrefine}, persistent skill libraries~\citep{wang2023voyager,wu2025evolver}, experiential memory~\citep{zhao2024expel,park2023generative,liu2025cer}, multi-agent coordination~\citep{wu2023autogen,du2024debate,qian2024chatdev}, in-context RL~\citep{laskin2023ad,chen2021dt,son2025dicp}, textual optimization that treats verbal feedback as a gradient-like signal over prompts or programs~\citep{yang2023opro,yuksekgonul2024textgrad}, and LLM-generated dense reward or critique~\citep{ma2024eureka,chen2024automanual,xie2025ctrl}. However, all these methods rely on a supervisory signal that is both \emph{rich} and \emph{cheap}, typically validated on benchmarks where per-trial cost is seconds to minutes (HotpotQA, AlfWorld, code unit tests). We extend verbal RL to the opposite regime: each trial costs GPU-hours of MD execution and yields only a single scalar at horizon end.

\subsection{Why Verbal RL for MDForge}
\label{app:why-verbal-rl}
Because $\theta$ is fixed, all learning is carried by the context rewrite $\mathcal{C}_t \mapsto \mathcal{C}_{t+1}$; the design of $\mathcal{C}_t$ is therefore the operative learning rule. No other update rule fits the regime: classical sparse-reward methods (reward shaping~\citep{ng1999shaping}, Bayesian optimization~\citep{shahriari2016taking}, hindsight relabeling~\citep{andrychowicz2017her}) presume a feature space that $\Pi$ does not admit; gradient-based RL~\citep{schulman2017ppo} demands orders of magnitude more rollouts; static few-shot prompting~\citep{brown2020fewshot} absorbs no trial signal at all. Verbal RL alone (i) treats the trial signal as text, so heterogeneous per-stage diagnostics fold back in without a fixed feature representation, (ii) keeps the policy a frozen LLM, so per-trial updates cost zero parameter passes, and (iii) keeps the action space open-ended.

\subsection{Process Supervision}
\label{app:related:process}
Process supervision densifies a sparse outcome signal by attaching intermediate scalar rewards to individual reasoning or decision steps. This idea spans modern language-model verification and test-time search~\citep{uesato2022process,cobbe2021gsm8k,lightman2023verify,wang2024mathshepherd,luo2024omegaprm,zhang2024restmcts,guo2025r1,snell2024scaling,zhang2025qwenprm,khalifa2025thinkprm,zheng2025processbench}, as well as classical reinforcement-learning densification~\citep{ng1999shaping,andrychowicz2017her,arjonamedina2019rudder,vezhnevets2017feudal,pathak2017curiosity,burda2019rnd,christiano2017preferences}. However, all these methods assume scalar feedback consumed via gradient-based parameter updates. We extend process supervision to the verbal, in-context regime: the densified signal is typed natural-language critique consumed without any parameter update.

\subsection{Multi-Agent Debate}
\label{app:related:debate}
A line of work casts inference-time reasoning as deliberation between multiple LLM instances, either as peers exchanging arguments to converge on a more reliable answer~\citep{du2024debate,liang2024debate} or as evaluators substituting for human judges~\citep{zheng2024judgelm,bai2022constitutional,lee2023rlaif}, with broader role-decomposition extending to multi-agent software- and task-execution frameworks~\citep{wu2023autogen,qian2024chatdev,li2023camel,ma2025autodata}. These methods treat experts as interchangeable reasoners differentiated only by prompt persona, and the quantity they produce is a single converged judgment over a shared query. The PRISM panel inside MDForge departs on both counts: each expert is anchored to a fixed physics subsystem (force field, sampling, analysis) with non-overlapping jurisdiction, and the output is a typed, subsystem-attributable critique that names which part of the pipeline to edit rather than a consensus verdict, with a slow cross-task loop reweighting per-expert reputations from their pre-trial vs.\ post-execution consistency.

%% file: appendix/method.tex
\section{Method Complement}
\label{app:method-details}

This appendix complements the high-level description of MDForge in \S\ref{sec:method} with the three technical pieces a reader needs to reproduce the system: the full PRISM pseudocode, a per-agent summary of the prompts that drive every LLM call together with the debate protocol and aggregator, and the reputation update. Verbatim prompts are released with the code.

\subsection{PRISM Pseudocode}
\label{app:method:pseudocode}

See Algorithm~\ref{alg:mdforge}.

\begin{algorithm}[h]
\small
\caption{\textbf{MDForge's verbal RL update.}}
\label{alg:mdforge}
\begin{algorithmic}[1]
\Require task $T$ (host + guest set); expert panel $E = \{e_{\mathrm{FF}}, e_{\mathrm{Samp}}, e_{\mathrm{Anal}}\}$; successful-trial budget $N$; failed-revision cap $M$; $K{=}4$ stages
\State $d_{\text{rec}} \gets \textsc{DesignDiscussion}(E, T)$
  \Comment{Phase 0: one-off, web-search enabled}
\State $\pi \gets \textsc{CodeAgent.Propose}(T, d_{\text{rec}})$
\State $n \gets 0$;\quad $m \gets 0$;\quad $H \gets \emptyset$;\quad $\pi^{*} \gets \bot$
\While{$n < N$ and $m < M$ and not converged}
  \State \Comment{\emph{Phase 1: static and mechanical screening}}
  \State $c_{L1} \gets \textsc{Layer1.Verify}(\pi)$
  \If{$c_{L1}.\text{label} = \textsc{Fail}$}
    \State $m \gets m + 1$;\quad $\pi \gets \textsc{CodeAgent.Revise}(\pi, [c_{L1}])$; \textbf{continue}
      \Comment{bypass panel}
  \EndIf
  \State $\pi \gets \textsc{Engineer.Debug}(\pi, T)$
    \Comment{mechanical fixes only; no methodological changes}
  \State \Comment{\emph{Phase 2: pre-execution verbal screening}}
  \State $(c_{\text{pre}}, S_{\text{pre}}) \gets \textsc{Aggregate}(\{e_i.\textsc{PreEval}(\pi)\}_{i})$
    \Comment{advisory}
  \State \Comment{\emph{Phase 3: engineer-led execution}}
  \State $D \gets \textsc{Engineer.Run}(\pi, T)$
    \Comment{debug + run stages $1..K$ in sandbox}
  \If{any required stage fails or no finite $\hat{\Delta G}$ is produced}
    \State $m \gets m + 1$;\quad $\pi \gets \textsc{CodeAgent.Revise}(\pi, [c_{L1}, c_{\text{pre}}, D])$; \textbf{continue}
  \EndIf
  \State $B \gets \textsc{MultiMoleculeBenchmark}(\pi, T)$
    \Comment{parallel over guest set}
  \State $(c_{\text{post}}, S_{\text{post}}) \gets \textsc{Aggregate}(\{e_i.\textsc{PostEval}(\pi, B)\}_{i})$
  \State \Comment{\emph{Phase 4: sparse outcome reward}}
  \State $r^{*} \gets -\,\textsc{MAE}(B)$;\quad $n \gets n + 1$;\quad $m \gets 0$
  \If{$c_{\text{post}}.\text{label} = \textsc{Pass}$ and $S_{\text{post}} \geq \theta_{\text{pass}}$}
    \State $\pi^{*} \gets \pi$; \textbf{break}
  \EndIf
  \State \Comment{\emph{Phase 5: in-context Code agent update}}
  \State $K_t \gets \{c_{L1}, c_{\text{pre}}, c_{\text{post}}\}$;\quad $H \gets H \cup \{D, B, r^{*}\}$
  \State $\mathcal{C}_t \gets \{T,\, \pi,\, K_t,\, H\}$
    \Comment{$K_t$ = critique set; $H$ = prior trial summaries}
  \State $\pi \gets \textsc{CodeAgent.Revise}(\pi, \mathcal{C}_t)$
  \State \Comment{\emph{Phase 6: per-task slow-loop reputation}}
  \State $\rho \gets \textsc{UpdateReputations}(\rho, c_{\text{pre}}.\text{votes}, c_{\text{post}}.\text{votes})$
\EndWhile
\State \Return $\pi^{*}$ if converged, else last runnable $\pi$
\end{algorithmic}
\end{algorithm}

\subsection{Multi-Agent Debate Protocol}
\label{app:method:panel}

We present the debate protocol and aggregator that turn per-expert votes into a single typed critique. The prompt of each expert agent is deferred in Appendix~\ref{app:prompts}. 
PRISM uses $J{=}3$ specialists with fixed jurisdictions: \emph{Force Field}, \emph{Sampling}, and \emph{Analysis}. The Analysis expert also covers restraints, standard-state correction, and thermodynamic-cycle algebra.

\paragraph{Shared expert output contract.}
Every expert returns the same JSON schema across design recommendation, pre-execution review, and benchmark post-execution review: a label $\ell_i \in \{\textsc{Pass}, \textsc{Uncertain}, \textsc{Fail}\}$, confidence $\kappa_i \in [0,1]$, a load-bearing \emph{strategic\_insight} field, a list of severity-scored concerns, and a short reasoning synthesis. Cross-domain commentary is allowed only when another subsystem directly affects the expert's jurisdiction.

\paragraph{Debate protocol.}
Given a fixed panel of $J$ experts, a single call to the panel runs two rounds. \textit{(i)} Round 1 (independent, parallel). Each expert receives the task description and either the pipeline source (pre-execution) or the pipeline plus its execution results (post-execution), and emits an independent vote in the shared schema. All $J$ experts run concurrently.
\textit{(ii)} Round 2 (cross-visibility, parallel). Each expert is shown the round-1 votes of the other $J{-}1$ experts and may revise its own vote in the same schema. All $J$ experts again run concurrently.

The same two-round protocol is used in three modes: design recommendation (Phase~0, before any pipeline exists), pre-execution review (\textsc{PreEval}, after the Code agent emits $\pi$ and the Engineer has removed mechanical faults), and benchmark post-execution review (\textsc{PostEval}, after the benchmark returns per-guest $\hat{\Delta G}$).

\paragraph{Aggregator $\mathcal{A}_\rho$.}
The aggregator collapses the round-2 votes into the single typed critique $c$ used by the Code agent. Each vote carries a label $\ell_i \in \{\textsc{Pass}, \textsc{Uncertain}, \textsc{Fail}\}$, a confidence $\kappa_i \in [0,1]$, and a free-text strategic insight. The aggregator scores each label as $s(\textsc{Pass}){=}1$, $s(\textsc{Uncertain}){=}0.5$, $s(\textsc{Fail}){=}0$, and computes the reputation- and confidence-weighted mean
\begin{equation}
S \;=\; \frac{\sum_i \rho_i\, \kappa_i\, s(\ell_i)}{\sum_i \rho_i\, \kappa_i},
\label{eq:aggregator-score}
\end{equation}
which is collapsed to a single panel label via two thresholds: $S \geq 0.7 \Rightarrow \textsc{Pass}$, $S \leq 0.3 \Rightarrow \textsc{Fail}$, otherwise $\textsc{Uncertain}$. The accompanying critique text is built by concatenating the per-expert strategic-insight and concern strings in reputation-weighted order; the full per-expert transcript is also retained for the Code agent's next-trial context.

\subsection{Reputation Update and Convergence}
\label{app:method:reputation}

Each expert $i$ carries a Beta posterior over its reliability, defined as the probability that its pre-execution vote is consistent with the post-execution outcome. The prior is uniform, $\mathrm{Beta}(\alpha_i{=}1,\beta_i{=}1)$. After each completed trial, the orchestrator inspects each expert's $(c_{\text{pre},t}, c_{\text{post},t})$ vote pair and applies a single Beta update per expert:
\begin{equation}
\begin{aligned}
\alpha_i \,\mathrel{+}&= \mathbf{1}\!\left[\text{expert } i \text{ consistent}\right],\\
\beta_i \,\mathrel{+}&= \mathbf{1}\!\left[\text{expert } i \text{ inconsistent}\right],
\end{aligned}
\label{eq:reputation-update}
\end{equation}
where ``consistent'' is the agreement between the pre-execution label of expert $i$ and the post-execution outcome of the trial (operationalized as the agreement between $i$'s pre-execution vote and the aggregator's post-execution label). The deterministic weight fed into the aggregator $\mathcal{A}_\rho$ is the posterior mean
\begin{equation}
\rho_i \;=\; \frac{\alpha_i}{\alpha_i + \beta_i},
\label{eq:reputation-mean}
\end{equation}
which is the maximum-likelihood estimate of reliability under the Beta--Bernoulli model. A Thompson-sampling variant draws $\tilde{\rho}_i \sim \mathrm{Beta}(\alpha_i, \beta_i)$ per trial; the deterministic posterior-mean form is the default reported in the main results.

\paragraph{Convergence within a task.}
Under the Beta--Bernoulli update of Eq.~\eqref{eq:reputation-update}, $\rho_i$ converges almost surely to expert $i$'s true reliability $p_i$ as the number of consistent/inconsistent observations grows, with $\mathrm{Var}(\rho_i) = \rho_i(1-\rho_i)/(\alpha_i+\beta_i+1) = \mathcal{O}(1/n_i)$. Within the $N{=}5$-successful-trial budget of a single task (a trial is counted only when the agent produces a runnable pipeline), $n_i \leq 5$ per expert, so the posterior mean has not converged to $p_i$; the update therefore functions as a soft prior that prevents the aggregator from equal-weighting an obviously-miscalibrated expert with reliable ones. The empirical $\rho_t$ trajectories per host are written to the per-task reputation log released with the code.

%% file: appendix/experimental_details.tex
\section{Experimental Details}
\label{app:implementation}

We provide the experimental details: LLM and MD configurations, hardware, trial protocol, and benchmark splits. Exact configurations and prompt texts are released with the code.

\paragraph{LLM configuration.}
We record, for each agent role in the MDForge loop (Code agent, Engineer agent, the $J{=}3$ panel experts, Layer-1 static verifier, and the aggregator $\mathcal{A}_\rho$), the model identifier, sampling temperature, maximum context, and retry policy. The same configuration is reused across all hosts; baselines are run on the same backbone with their own prompt templates. We use the Claude Opus 4.7 as the default backbone. 

\paragraph{Hardware and compute budget.}
Each guest evaluation runs in $\approx 2$ GPU-hours on a single A40 node. The verbal RL loop continues until $N{=}5$ successful trials accumulate per host (a trial is counted when the agent produces a runnable pipeline; attempts that abort at Layer-1, crash the Engineer, or are abandoned by the agent do not count toward $N$), so the total round count per host can exceed five.

\paragraph{Trial protocol.}
A trial begins with the Code agent emitting a pipeline and ends either when the pipeline aborts at a stage boundary or when it completes the analysis stage and returns a $\hat{\Delta G}$ per training guest. \emph{Coding success} counts a trial whose emitted code is runnable, regardless of whether it later crashes or does not converge inside MD. The \emph{best pipeline} reported in Table~\ref{tab:main} is selected, among trials that complete the production stage on all four training guests, as the one with the highest training-set Kendall $\tau$; the same selection rule is applied to all methods.

\paragraph{Benchmark splits.}
For each of the three hosts (CB[7], OAH, CBClip), the guest set is split into four training guests (visible to the verbal RL feedback loop) and the remaining guests as a held-out test set. The training quadruple is chosen to span the experimental $\Delta G$ range of each host and is fixed across all methods to keep comparisons aligned. See Table~\ref{tab:benchmark-splits} for details. 

\begin{table}[t]
  \centering
  \small
  \setlength{\tabcolsep}{2pt}
  \begin{tabular}{@{}l c l l@{}}
    \toprule
    System & $N$ total & Train indices & Held-out indices \\
    \midrule
    SAMPL4 CB[7] & 14 & 1, 5, 8, 10 & \begin{tabular}[t]{@{}l@{}}0, 2, 3, 4, 6,\\7, 9, 11, 12, 13\end{tabular} \\
    SAMPL4 OAH & 9 & 0, 3, 5, 8 & 1, 2, 4, 6, 7 \\
    SAMPL5 CBClip & 10 & 0, 2, 3, 4 & \begin{tabular}[t]{@{}l@{}}1, 5, 6, 7,\\8, 9\end{tabular} \\
    \bottomrule
  \end{tabular}
  \caption{\textbf{Benchmark train/test splits.} Indices are zero-based within each host's guest list.}
  \label{tab:benchmark-splits}
\end{table}

%% file: appendix/additional_results.tex
\section{Uncertainty on the Held-Out Rank Correlations}
\label{app:add:uncertainty}

The held-out sets are small ($n_{\mathrm{held}} = 10$, $5$, and $6$ for CB[7], OAH, and CBClip), so a rank correlation can move on a few pairwise reorderings. Table~\ref{tab:uncertainty} reports two robustness checks on the selected trial: a nonparametric bootstrap $95\%$ confidence interval, and a leave-one-out (LOO) range obtained by dropping each held-out guest in turn and recomputing $\tau$.

\begin{table}[!h]
  \centering
  \scriptsize
  \setlength{\tabcolsep}{3.2pt}
  \renewcommand{\arraystretch}{1.10}
  \begin{tabular}{@{}llccc@{}}
    \toprule
    Host & Method & $\tau$ & LOO range & Bootstrap 95\% CI \\
    \midrule
    \multirow{3}{*}{CB[7]}
      & Step-level  & 0.16 & $[-0.06,\ 0.61]$ & $[-0.65,\ 1.00]$ \\
      & Trial-level & 0.24 & $[-0.06,\ 0.44]$ & $[-0.60,\ 1.00]$ \\
      & \textbf{MDForge} & \textbf{0.56} & $\mathbf{[0.33,\ 0.78]}$ & $[-0.18,\ 1.00]$ \\
    \midrule
    \multirow{3}{*}{OAH}
      & Step-level  & 0.20 & $[-0.33,\ 0.33]$ & $[-1.00,\ 1.00]$ \\
      & Trial-level & 0.20 & $[-0.33,\ 0.33]$ & $[-1.00,\ 1.00]$ \\
      & \textbf{MDForge} & \textbf{0.20} & $\mathbf{[0.00,\ 0.67]}$ & $[-1.00,\ 1.00]$ \\
    \midrule
    \multirow{3}{*}{CBClip}
      & Step-level  & 0.07 & $[-0.40,\ 0.40]$ & $[-1.00,\ 1.00]$ \\
      & Trial-level & 0.20 & $[-0.20,\ 0.40]$ & $[-0.67,\ 1.00]$ \\
      & \textbf{MDForge} & \textbf{0.47} & $\mathbf{[0.20,\ 0.60]}$ & $[-0.33,\ 1.00]$ \\
    \bottomrule
  \end{tabular}
  \caption{\textbf{Uncertainty on held-out Kendall $\tau$.} Point estimate, leave-one-out range (drop one held-out guest, recompute), and nonparametric bootstrap $95\%$ CI. At $n_{\mathrm{held}} = 5$--$10$ the bootstrap intervals are uninformative by construction; the LOO ranges carry the signal.}
  \label{tab:uncertainty}
\end{table}

The bootstrap intervals span most of $[-1, 1]$ at this sample size and are essentially uninformative; we report them for completeness. The LOO range is the sharper instrument here, because it perturbs the evaluation set exactly one guest at a time, which is the failure mode of concern. Under LOO the ordering is stable: on CB[7] the entire MDForge range (lower bound $0.33$) lies above both baselines' point estimates ($0.24$, $0.16$), and on CBClip the MDForge lower bound ($0.20$) meets or exceeds them ($0.20$, $0.07$). The baselines are also less stable than MDForge, dipping below zero on CB[7] and OAH, i.e.\ dropping a single guest can flip them to anti-correlated, whereas MDForge stays non-negative on all three hosts. No single held-out guest is responsible for the reported gain.

\section{Reputation-Weighted vs.\ Equal-Weighted Aggregation}

Table~\ref{tab:reputation} isolates the reputation loop of \S\ref{sec:method:writer} by replacing $\rho_t$ with uniform expert weights, holding the panel, the stage diagnostics, and the code agent fixed. Reputation weighting gives a small but consistent gain ($0.40 \pm 0.14$ vs.\ $0.33 \pm 0.11$ on CB[7]). We are explicit that within a five-trial budget it acts as a stabilisation component rather than the principal driver: the dominant contribution comes from the typed per-stage diagnostics and subsystem-scoped critique, whose removal costs far more (Table~\ref{tab:main}).

\begin{table}[!h]
  \centering
  \scriptsize
  \setlength{\tabcolsep}{3.2pt}
  \renewcommand{\arraystretch}{1.10}
  \begin{tabular}{@{}lccccccc@{}}
    \toprule
    \multirow{2}{*}{Aggregation} & \multicolumn{5}{c}{Trial} & \multirow{2}{*}{Mean $\pm$ std} \\
    \cmidrule(lr){2-6}
    & 1 & 2 & 3 & 4 & 5 & \\
    \midrule
    Equal weighting & 0.33 & 0.16 & 0.33 & 0.42 & 0.42 & $0.33 \pm 0.11$ \\
    \textbf{Reputation} & 0.20 & 0.33 & 0.42 & 0.51 & 0.56 & $\mathbf{0.40 \pm 0.14}$ \\
    \bottomrule
  \end{tabular}
  \caption{\textbf{Reputation-weighted vs.\ equal-weighted expert aggregation on CB[7].} Per-trial held-out Kendall $\tau$ over $N{=}5$ successful trials, with all other components held fixed.}
  \label{tab:reputation}
\end{table}

%% file: appendix/prompt.tex
\section{Agent Prompts}
\label{app:prompts}

We present the system prompts that define each agent's role inside MDForge. Three domain experts (force-field, sampling, analysis) act as peer co-designers and reviewers, while a single code-writing agent, split between the Pipeline Writer and the Pipeline Engineer, is responsible for producing and debugging the four-stage Python pipeline. Each panel below distills the operative content of the corresponding system prompt; the full prompts are released with the code.

\newtcolorbox{promptpanel}[2][]{
  enhanced,
  colback=#2!4,
  colframe=#2!55!black,
  colbacktitle=#2!18,
  coltitle=black,
  fonttitle=\bfseries,
  title={#1},
  boxrule=0.5pt,
  titlerule=0.3pt,
  arc=2pt,
  left=6pt,right=6pt,top=4pt,bottom=4pt,
  before skip=6pt, after skip=6pt,
}

\begin{figure*}[!t]
\begin{promptpanel}[Force-Field Expert: system prompt summary]{NavyBlue}
\textbf{Persona.} A senior computational chemist with 10+ years on small-molecule force-field development and on the parameterization of SAMPL3--9 host-guest benchmarks. The agent participates as a peer in pipeline design, not as a narrow gatekeeper.

\smallskip
\textbf{Primary domain.}
\begin{itemize}[leftmargin=*,itemsep=1pt,topsep=1pt]
  \item \emph{Guest force fields.} GAFF/GAFF2+AM1-BCC as SAMPL default, with known $\sim$1\,kcal/mol over-binding bias on cation-$\pi$ contacts; OpenFF (Sage/Parsley) for more robust SMIRNOFF typing on non-standard cations; CGenFF only when the host is also CHARMM-parameterized. Aware of the canonical antechamber failure where tertiary ammonium nitrogen is mistyped as \texttt{nz} (sp$^2$) instead of \texttt{n4}.
  \item \emph{Host force fields.} CB7 is parameterized with the same family as the guest. Charge provenance of the SAMPL-distributed \texttt{cb7.mol2} is treated as suspect; regenerating with AM1-BCC for consistency is the most defensible option.
  \item \emph{Water and ions.} TIP3P as the SAMPL baseline, with $\sim$0.5--1\,kcal/mol bias toward less-negative $\Delta G$ around cationic ammoniums. OPC or TIP4P-Ew paired with matched ion sets (Joung-Cheatham for TIP3P/SPC; OPC-trained ions for OPC) for improved cation hydration. Never mix water-specific ions with the wrong water model.
  \item \emph{File consistency.} Atom-type assignments must agree between mol2, frcmod and tleap; \texttt{parmchk2} must run after any atom-type rewriting; \texttt{tleap.log} should be scanned for ``Could not find$\dots$'' warnings.
\end{itemize}

\smallskip
\textbf{Co-design behaviour.} Before enumerating concerns, the agent commits to a strategic position: for the specific chemical class at hand, what force-field choice would it make and why. It then audits whether the Writer's choice is defensible by that standard. In post-eval, it reads anomalous energy components and the reported $\Delta G$ through the lens of class-specific systematic biases (e.g., GAFF2+AM1-BCC+TIP3P should land $\sim$1--2\,kcal/mol less negative than the experimental reference for CB7-cation systems). Cross-domain remarks on sampling, restraints or analysis are welcome when they affect whether force-field concerns are even detectable.

\smallskip
\textbf{Operating modes.} The same agent is invoked in four modes: (A) design recommendation before any pipeline exists, (B) pre-eval critique of a proposed design, (C) post-eval interpretation of a single-molecule run, and (D) post-eval over a multi-molecule benchmark, where it proposes the specific, surgical pipeline edit (named file, region and change) that would most improve next-iteration MAE.

\smallskip
\textbf{Output contract.} Each invocation returns a single JSON object with fields \texttt{label} $\in\{$\texttt{pass},\texttt{fail},\texttt{uncertain}$\}$, \texttt{confidence}, a load-bearing \texttt{strategic\_insight}, a list of \texttt{concerns} (each with severity and suggested focus), and a final \texttt{reasoning} synthesis. The \texttt{strategic\_insight} is asked to be method-level rather than parameter-level, system-aware, comparative across alternatives, and literature-grounded.
\end{promptpanel}
\end{figure*}

\begin{figure*}[!t]
\begin{promptpanel}[Sampling Expert: system prompt summary]{OliveGreen}
\textbf{Persona.} A senior molecular-simulation methodologist with 10+ years of experience designing alchemical, APR and umbrella-sampling protocols for binding free energies. Calibrated on hundreds of SAMPL-style benchmarks and able to recognize an under-sampled protocol from the $\lambda$ schedule alone.

\smallskip
\textbf{Primary domain.}
\begin{itemize}[leftmargin=*,itemsep=1pt,topsep=1pt]
  \item \emph{Strategy choice.} Two main families for absolute binding: Attach-Pull-Release (APR, Henriksen-Gilson; the SAMPL CB7 standard; typically 15--25 windows, 1--5\,ns each) and alchemical absolute binding (less common on CB7, with known pitfalls around PME+decoupling in openmmtools, softcore LJ at $\alpha{=}0.5$, electrostatics-first $\lambda$ schedules, and dense LJ-endpoint spacing).
  \item \emph{Integrator and timestep.} \texttt{LangevinMiddleIntegrator} as modern default; 2\,fs with HBonds constraints, 4\,fs with HBonds+HMR; 1/ps friction. 5\,fs HMR is aggressive and requires validation.
  \item \emph{Equilibration.} Minimize, then NVT heating (100--500\,ps), then NPT density (0.5--2\,ns; longer for charged guests). A pipeline that skips NPT is a red flag.
  \item \emph{Replicates and seeds.} For CB7-class systems a single long replicate is often acceptable; each replicate must be seeded independently. $\geq 3$ replicates is preferred but rarely fits the budget.
  \item \emph{Hardware and wall-clock.} OpenMM CUDA or \texttt{pmemd.cuda}, never silently CPU. The agent is explicitly briefed on a 2-hour cap on the production stage and is told to surface the trade-off (``24 windows $\times$ 2\,ns is defensible but exceeds budget; reduce to 12 or accept the violation'') rather than demand the impossible.
\end{itemize}

\smallskip
\textbf{Co-design behaviour.} The \texttt{strategic\_insight} field is required to answer three questions for the specific system class: (i) what sampling strategy is best practice (e.g., APR with Henriksen-Gilson corrections for CB7), (ii) is the Writer's choice defensible, and if it took the less-common alchemical route, why might that be reasonable, and (iii) what is the largest sampling-side risk to the reported $\Delta G$ under the chosen design and wall-clock budget. In post-eval, the agent reads wall-clock used vs.\ designed, integrator stability evidence, T/P/density drift, per-window dwell time, replicate scatter, autocorrelation, and MBAR overlap; cross-domain remarks on force-field or restraint choices are welcome when they affect sampling sufficiency.

\smallskip
\textbf{Operating modes \& output contract.} As for the force-field expert, the same JSON schema is emitted in all four modes (design recommendation, pre-eval, single-molecule post-eval, multi-molecule benchmark post-eval). In Mode~D the agent is asked to name a surgical pipeline edit (specific file, region, change) that should improve next-iteration MAE.
\end{promptpanel}
\end{figure*}

\begin{figure*}[!t]
\begin{promptpanel}[Analysis Expert: system prompt summary]{Plum}
\textbf{Persona.} A senior simulation methodologist whose specialty is statistical-mechanics estimators for free-energy calculations and the thermodynamic-cycle algebra around restraint application and release. Has implemented MBAR/BAR/TI from scratch, derived the Boresch standard-state correction from the Gaussian partition function, and spent years separating ``the calculation didn't converge'' from ``the estimator was wrong'' from ``the restraint correction has the wrong sign''.

\smallskip
\textbf{Primary domain.}
\begin{itemize}[leftmargin=*,itemsep=1pt,topsep=1pt]
  \item \emph{Restraint design.} APR-style (1 distance along the host symmetry axis; analytic release; standard for CB7) versus Boresch (6-DOF, closed-form correction, overkill for symmetric hosts). Anchors are rigid heavy atoms (host ring carbon or carbonyl centroid; guest bridgehead or ammonium N). Sane CB7 force constants: $k_r{=}5\text{--}20$\,kcal/mol/\AA$^2$, Boresch angles/torsions 50--200\,kcal/mol/rad$^2$.
  \item \emph{Standard-state correction.} The agent is given the harmonic well integral $V_\text{well}{=}(2\pi kT/k)^{3/2}$ and $\Delta G_\text{release}{=}-kT\ln(V_\text{std}/V_\text{well})$ with $V_\text{std}{=}1660$\,\AA$^3$, and is explicitly warned that sign errors, leg misplacement, or a missing $-RT\ln(n_\text{sym})$ symmetry factor are the most common cause of $\Delta G_\text{bind}$ off by 5--15\,kcal/mol.
  \item \emph{Estimator choice.} MBAR by default for absolute binding via decoupling; BAR/TI when only adjacent pairs or $\langle\partial H/\partial\lambda\rangle$ are available; the Henriksen-Gilson three-term decomposition for APR; MM-PB/GBSA only as a rough first pass.
  \item \emph{$\lambda$ schedule.} Electrostatics-first, soft-core LJ with $\alpha{=}0.5$, denser LJ spacing near the endpoint, 8--12 electrostatics windows plus 12--15 LJ windows, MBAR overlap $\geq 0.03$ between adjacent states.
  \item \emph{Uncertainty.} Three layers: within-replicate (pymbar bootstrap/block-jackknife), across-replicate scatter, and systematic biases (force field, restraint correction, sampling); reporting $\pm 0.1$\,kcal/mol on a CB7 system is treated as a smell.
\end{itemize}

\smallskip
\textbf{Co-design behaviour.} The \texttt{strategic\_insight} field is asked to answer: which estimator and $\lambda$ schedule are right for the specific system, is the thermodynamic cycle algebra correctly implemented \emph{in code} (term-by-term sign check), and what is the biggest analysis-side risk to the reported $\Delta G$. The agent is given a numerical reasonableness band for CB7-adamantylammonium (literature $\Delta G\approx-14$\,kcal/mol; GAFF2+AM1-BCC+TIP3P should land in $[-14,-10]$; values outside $[-18,-8]$ have something wrong).

\smallskip
\textbf{Operating modes \& output contract.} As for the other experts: four-mode invocation and the shared JSON schema with \texttt{label}, \texttt{confidence}, \texttt{strategic\_insight}, \texttt{concerns}, and \texttt{reasoning}.
\end{promptpanel}
\end{figure*}

\begin{figure*}[!t]
\begin{promptpanel}[Pipeline Writer: system prompt summary]{BurntOrange}

\textbf{Pipeline Writer (code-authoring agent).}

\smallskip
\emph{Mandate.} Author a complete, runnable MD pipeline that computes the binding free energy of the host-guest pair specified by the user, by writing Python code from scratch as a sequence of $K{=}4$ sequential stages. Docking is out of scope: a bound-complex \texttt{complex.pdb} is pre-staged in the working directory and stage~01 reads it directly.

\smallskip
\emph{Step 0: literature reconnaissance.} Before writing a single line of code, the agent is required to use \texttt{WebSearch} and \texttt{WebFetch} for at least three queries combining the host class with terms such as ``binding free energy method'', ``SAMPL benchmark'', ``attach-pull-release'' and ``alchemical absolute binding'', and to fetch at least one methods paper. The \texttt{RATIONALE} paragraph must name the chosen method, cite at least one published reference for the choice on this host class, and explain why the method fits the system's physics. Methodological choices without a literature citation are not acceptable.

\smallskip
\emph{Four-stage pipeline.} (1) \texttt{01\_prep.py}: read \texttt{complex.pdb}, pick force field/charge model/water, parameterize the guest via \texttt{antechamber}+\texttt{parmchk2}, build the solvated topology in a single \texttt{tleap} call, then minimize. (2) \texttt{02\_equilibrate.py}: short NVT heating + NPT density, reporting T/P/density traces and drift. (3) \texttt{03\_production.py}: the expensive sampling stage, reading per-window sampling length from the \texttt{MDFORGE\_PRODUCTION\_NS\_PER\_WINDOW} environment variable. (4) \texttt{04\_analysis.py}: MBAR/TI/WHAM/BAR with restraint standard-state correction and $-RT\ln(n_\text{sym})$ symmetry correction, populating \texttt{delta\_g\_kcal\_per\_mol} with an uncertainty.

\smallskip
\emph{Molecule-agnostic invariant.} A single pipeline must run on every (host, guest) task with only the input files changing. The agent is forbidden from hardcoding guest names, atomic charges, symmetry numbers, or pH, and must read those from \texttt{task\_metadata.json}; only the canonical filenames \texttt{guest.mol2}, \texttt{host.mol2}, \texttt{complex.pdb} may appear in code.

\smallskip
\emph{Output contract.} The reply must follow an exact \texttt{RATIONALE} / \texttt{ENTRY} / \texttt{FILE} block structure that the harness parses mechanically; deviations cause an automatic Layer-1 failure. Each stage must write \texttt{stage\_NN\_result.json} on exit (even on failure) with \texttt{status}, \texttt{wall\_time\_seconds}, \texttt{delta\_g\_kcal\_per\_mol}, \texttt{convergence\_flags}, \texttt{energy\_components}, \texttt{diagnostics}, and a \texttt{writer\_notes} field that the verifier experts read.

\smallskip
\emph{Engineering pitfalls embedded in the prompt.} The Writer is briefed on environment-specific failure modes: openmmtools' PME-with-decoupled-electrostatics incompatibility (must set \texttt{annihilate\_electrostatics=True}), antechamber's \texttt{nz} vs.\ \texttt{n4} mis-typing on protonated tertiary amines (with the explicit warning that AM1-BCC charges on ammonium N are physically negative, so naive ``$N{>}0$'' sanity checks must not be inserted), \texttt{tleap} sourcing order (small-molecule FF before water leaprc), CB7 host-charge provenance, the requirement that result files be written before re-raising on failure, hard wall-clock budgeting for stage~03, mandatory GPU use, and a single-GPU-per-molecule invariant (no \texttt{ProcessPoolExecutor} across GPUs inside one pipeline).

\end{promptpanel}
\end{figure*}

\begin{figure*}[!t]
\begin{promptpanel}[Pipeline Engineer: system prompt summary]{BurntOrange}

\textbf{Pipeline Engineer (debug-and-run agent).}

\smallskip
\emph{Mandate.} Given the four stage files the Writer just emitted, debug, iterate, and run the pipeline end-to-end in the sandbox until stage~04 outputs a finite $\Delta G$. The session ends with a non-null \texttt{delta\_g\_kcal\_per\_mol} or it is considered failed; the next trial inherits worse context if it fails.

\smallskip
\emph{Execution discipline.} Stages are run sequentially with synchronous blocking \texttt{Bash} calls (\texttt{timeout} set generously). After each run the agent inspects \texttt{stage\_NN\_result.json} before proceeding. Stage~03 must not be re-run once it has succeeded: if stage~04 then crashes, only stage~04 is re-run against the existing per-window energies.

\smallskip
\emph{Hard rules (allowed vs.\ forbidden edits).} The Engineer may make mechanical fixes (switching \texttt{antechamber -c bcc}$\to$\texttt{-c rc} to skip a hang, removing a wrong sanity check, fixing \texttt{tleap} sourcing order, adding \texttt{os.makedirs(..., exist\_ok=True)}, catching format exceptions). It is forbidden from (i) changing methodological choices (APR$\to$DDM, GAFF2$\to$OPLS, TIP3P$\to$OPC, reducing production length), (ii) gaming QC gates to manufacture success (silencing flags, zeroing uncertainty, lowering thresholds), (iii) introducing molecule-specific hardcoding, or (iv) bypassing \texttt{MDFORGE\_PRODUCTION\_NS\_PER\_WINDOW} to make $\Delta G$ look better in the debug pass.

\smallskip
\emph{Forbidden tools.} The Engineer runs in a single, non-resumable session. \texttt{ScheduleWakeup}, \texttt{Monitor}, \texttt{ToolSearch}, and \texttt{run\_in\_background:\,True} are all explicitly disabled; the only available tools are \texttt{Read}, \texttt{Write}, \texttt{Edit}, \texttt{Bash}. Long-running stages are handled by blocking \texttt{Bash} with a large timeout rather than backgrounding.

\smallskip
\emph{Exit conditions.} Success: \texttt{stage\_04\_result.json} carries a finite numeric $\Delta G$ and all earlier stages completed cleanly. Time budget: $\sim$60 minutes wall-clock for the whole debug session; the agent wraps up gracefully if running short. Stuck: if the same kind of fix has been tried twice on the same stage and the same error keeps coming back, the agent stops and documents the blocker for the next trial's Writer to handle.

\end{promptpanel}
\end{figure*}